\def\year{2019}\relax
\documentclass[letterpaper]{article} 
\usepackage{aaai19}  
\usepackage{times}  
\usepackage{helvet}  
\usepackage{courier}  
\usepackage{url}  
\usepackage{graphicx}  
\usepackage{latexsym}
\usepackage{graphicx}
\usepackage{amsmath}
\usepackage{amssymb}
\usepackage{color}
\usepackage{booktabs}
\usepackage{array}
\usepackage{bm}
\usepackage{multirow}
\newcolumntype{L}{>{\centering\arraybackslash}m{3cm}}
\newcommand{\STAB}[1]{\begin{tabular}{@{}c@{}}#1\end{tabular}}
\newcommand{\newcite}[1]{\citeauthor{#1} \shortcite{#1}}
\DeclareMathOperator*{\argmax}{argmax}

\begin{document}
%
\author{
Parag Jain,
Abhijit Mishra,
Amar Prakash Azad,
Karthik Sankaranarayanan\\
IBM Research\\
\{pajain34,abhijimi,amarazad,kartsank\}@in.ibm.com}
\title{Unsupervised Controllable Text Formalization}
\maketitle
\date{}
\begin{abstract}
We propose a novel framework for controllable natural language transformation. Realizing that the requirement of parallel corpus is practically unsustainable for controllable generation tasks, an unsupervised training scheme is introduced. The crux of the framework is a deep neural encoder-decoder that is reinforced with text-transformation knowledge through auxiliary modules (called \emph{scorers}). These scorers, based on off-the-shelf language processing tools,  decide the learning scheme of the encoder-decoder based on its actions. We apply this framework for the text-transformation task of formalizing an input text by improving its readability grade; the degree of required formalization can be controlled by the user at run-time. Experiments on public datasets demonstrate the efficacy of our model towards: (a) transforming a given text to a more formal style, and (b) varying the amount of formalness in the output text based on the specified input control. Our code and datasets are released for academic use.
\end{abstract}
\section{Introduction}
\label{sec:intro}
Automatic text style-transformation is one of the key goals of \emph{text-to-text} natural language generation (NLG) research and most existing systems for such tasks are either supervised (\textit{e.g.,} variants of \emph{Seq2Seq} neural models \cite{sutskever2014sequence,bahdanau2014neural}, or Statistical Machine Translation models \cite{koehn2009statistical}) or  template/rule based \cite{gatt2009simplenlg}. Supervised NLG requires large-scale parallel corpora for training, which is a major impediment in scaling to diverse use-cases. For example, in the context of commercial dialog systems alone, there are several scenarios where a system's answer (which may be coming from a database \cite{N18-2098}) needs to be transformed either for its \emph{tone} (politeness, excitedness, \textit{etc.}), or its level of \emph{formality} (casual, formal, \textit{etc.} based on the user's personality), or for its \emph{complexity} (simplifying linguistic or domain-specific terminology such as in legal or medical domains). As such requirements and use-cases keep growing, it is practically unsustainable to obtain large scale parallel corpora for each such text transformation task. 

From a scientific perspective, a supervised treatment of all such tasks using several parallel corpora seeks to learn both the language transformation (while preserving semantics), as well as the style transformation, simultaneously for each task. We make 3 key observations with respect to this: (i) since the preservation of language semantics is necessary for transformation, whereas only the attribute or style of the text needs to be changed, it should be possible to decouple these two aspects, (ii) it should be cheaper to independently \emph{verify} these aspects at the output (with well-understood NLP techniques) than to \emph{specify} the required transformation for each input text with its output example, and (iii) it should be possible to \emph{control} the degree or magnitude of the intended attribute (readability level, politeness level, \textit{etc.}) required at the output. These observations motivate us to seek an unsupervised approach to such a gamut of text transformation tasks and underlie the proposed NLG framework.
\begin{figure}[t]
\centering
\footnotesize
\includegraphics[width=.70\columnwidth, keepaspectratio]{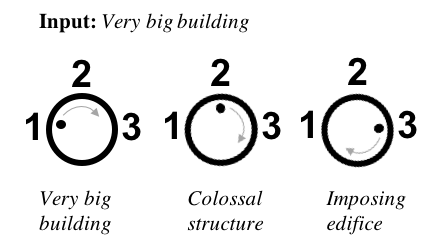}
\caption{Controllable formalization showing an anecdotal snippet and its expected variations w.r.t. input controls}
\label{fig:example}
\end{figure}

Our proposed framework relies only on unlabeled texts for initialization and an ensemble of \textit{off-the-shelf} language processing modules. We test our framework for the task of \emph{formalizing  the input text}, where the readability of the output text is improved while preserving its meaning (refer Figure \ref{fig:example} for an example). 
The degree of formalization required may be decided by the user during run-time and is provided as a control input to the system. This task is chosen due to its relevance in many NLG applications such as formal conversation generation, email response composition, or summary document generation in compliance and regulatory domains \textit{etc}. Moreover, such a standalone system can provide assistance to  professional writers, the same way Computer Assisted Translation (CAT) systems currently assist human translators. This paves the way for cost- and time-effective solutions for textual content creation.

Our framework is based on an encoder-decoder module \cite{bahdanau2014neural}, which is pre-trained with unlabeled texts. \textbf{The decoder additionally takes user-specified control as input}. Further, knowledge of the required text-transformation is acquired through auxiliary modules (called \emph{scorers}) which decide the learning scheme of the encoder-decoder based on its actions. These scorers are based on readily available natural language processing (NLP) tools which can produce scores indicating: (a) how formal the generated text is, (b)  whether the generated text is fluent, and most importantly, (c) whether the generated text carries similar  semantics as the input. This framework is trained in multiple iterations, where each iteration is comprised of two phases of \textbf{(i) exploration} and \textbf{(ii) exploitation}. In the exploration phase, the decoder randomly samples candidate texts for given inputs, and with the help of scorers, automatically produces training data for controllable generation. In the exploitation phase, the encoder-decoder is retrained with the examples thus generated. 

For experiments, we prepare a mixture of unlabeled informal texts with low readability grade. Our NLP tools measure readability, adequacy and fluency. With this setup, we observe that (a) the system generated transformed texts are more formal than the input, and (b) their degree of formalness conforms to the input controls provided. The efficacy of our system is demonstrated through both qualitative and quantitative evaluations, in terms of human judgments and various NLG evaluation metrics. We also show the system's effectiveness for another relevant task of text complexification (reversed simplification). The source code and dataset are publicly available. 
\section{Related Work}
\label{sec:related}
Unsupervised NLG has always been more challenging due to the fact that: (i) the output space is more complex and structured, making unsupervised learning more difficult, and (ii) metrics for evaluating NLG systems without reference output text are elusive. 
Recently, \newcite{artetxe2017unsupervised} and \newcite{lample2017unsupervised} have proposed architectures for unsupervised language translation with unsupervised autoencoder based lexicon induction techniques. This approach primarily focuses on cross-lingual transformation and requires multiple unlabeled corpora from different languages. It can not trivially be extended to our setup to achieve a controllable text-transformation goal within a single language, and further there is no notion of control in language translation. A notable work by \newcite{hu2017toward} discusses controllable generation; the system takes control parameters like sentiment, tense \textit{etc.}, and generates random sentences conforming to the controls. 
However, this system, unlike ours, does not transform a given input text.

The work that is most relevant to ours is by \cite{mueller2017sequence} which jointly trains a VAE and an outcome prediction module to correct an input such that the output has a higher expected outcome. We use this work for comparison by configuring their system to perform formal style transformation. This model, however, does not take as input an external control parameter which ours does.

\begin{figure*}[t]
\centering
\includegraphics[width=.70\textwidth, keepaspectratio]{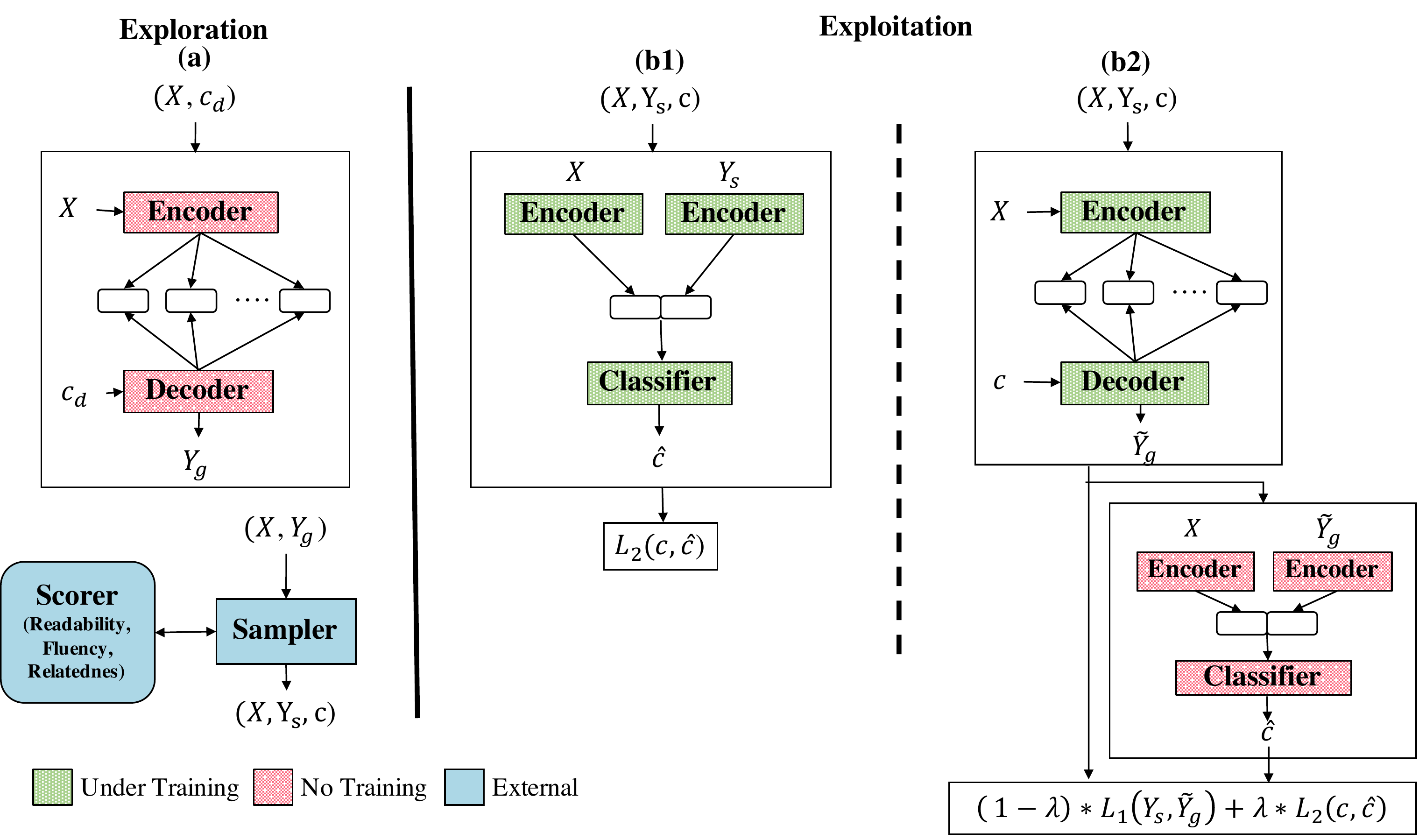}
\caption{System architecture and phases of training}
\label{fig:architecture}
\end{figure*}
Some other relevant works are on sentiment and attribute based unsupervised style transfer \cite{ficler2017controlling,shen2017style,shen2017style}, semi-supervised transfer through back translation using a translation corpora \cite{prabhumoye2018style}, formal-informal text classification using linguistic feature \cite{seika2012classify}, politeness analysis  \cite{danescuniculescumizil-EtAl:2013:ACL2013}, polite-conversation generation \cite{niu2018polite} using encoder-decoder models, but these do not perform \textbf{controllable} text-transformation. Similarly other relevant generation frameworks for formal-text generation by \newcite{sheikha2011generation} and paraphrase generation \cite{wubben2010paraphrase,prakash2016neural} are either template based or supervised, and are not controllable. Language generation systems such as \newcite{li2017paraphrase,Yu2017SeqGANSG} which incorporate NLP based scorers are unsupervised but suffer from convergence problems while training, as also  pointed out by \newcite{hu2017toward}. To the best of our knowledge, our work is the first to approach the task of \textit{unsupervised controllable text transformation}.

\section{System Overview}
\label{sec:training}
Our framework is designed to take text (or sentence, at this moment) and a set of control parameters. At this moment we consider only one input control \textit{i.e.}, degree of formalization. For such a task, a neural encoder-decoder is a natural choice. In our setting, the encoder-decoder takes a control value apart from the input text. To train this module without supervision, additional components are employed. Figure \ref{fig:architecture} depicts an overview of the system and the learning scheme.
\subsection{Encoder and Controllable-Decoder}
\label{subsec:encdec}
Our framework builds on the encoder-attend-decoder architecture of \cite{bahdanau2014neural}. The working principle of this is well-known, and is thus skipped for brevity. A vital change that we apply to the traditional architecture is the inclusion of an additional input to the decoder - the \text{control input}. The control input is passed through an embedding layer of $d$ dimensions.

Given an input sentence $\bm{X}$ comprising $N$ words, and a $d$ dimensional control vector $\bm{c}$, the encoder first produces a hidden representation $\bm{H}$, using stacked layers of GRUs \cite{cho2014properties}. The decoder generates one word at each time step, by: (a) \emph{attending} to the hidden representation  building a context vector, (b) concatenating the context vector with the control vector and the current hidden state of the decoder, and (c) producing a conditional probability distribution from the resultant output through a non-linear function. To summarize, for the $M$ word output sentence $\bm{Y}=(\bm{y_1},\bm{y_2},\cdots, \bm{y_{M}})$, the conditional probability term for each word $y_i$ can be computed as,
\begin{equation}
\begin{split}
p (\bm{y_{i}} | \{\bm{y_1},\bm{y_2},\cdots,\bm{y_{i-1}}\}, \bm{z_{i}},\bm{c}) \\
= g(\bm{E}\bm{y_{i-1}},\bm{c},\bm{s_{i}},\bm{z_{i}})
\label{eq:cond}
\end{split}
\end{equation}
where $g$ is the \texttt{softmax} function that outputs the probability distribution of $\bm{y_{i}}$; $\bm{E}\bm{y_{i-1}}$ is the embedding of the previously generated word; $\bm{c}$ is the control vector;  $\bm{z_{i}}$ is the vector obtained by attending over $\bm{H}$; $\bm{s_{i}}$ is the hidden state of the decoder GRU at time-step $i$.

\subsection{Sampler and Scorer}
\label{subsec:sampler}
To train the encoder-decoder framework in an unsupervised manner (\textit{i.e.} without true labels $\bm{Y}$ being available), instead of taking the most likely output of the decoder, the output probability distribution is passed through a \textit{sampler} that samples $K$ different samples from the decoder. Sampling is necessary for the \textbf{exploration} step in the training process (discussed later in Section \ref{subsec:explore}). Given an input text $\bm{X}$, the decoder can produce output $\bm{Y_g}$. The role of the sampler is to take $\bm{Y_g}$ and produce $K$ variants of $\bm{Y_g}$ by applying slight variations.  Once $K$ samples are obtained, the best sample ($\bm{Y_{s}}$) that maximizes the weighted score (produced by the scorer module), is retained for further processing. 
\begin{equation} \small
\bm{Y_{s}} = \argmax_{\bm{Y}} \{G(\bm{X},\bm{Y}) | \bm{Y} \in \{\bm{Y_{g}},Sample_{K}(\bm{Y_g})\}\}
\end{equation}
where $G(.)$ is the scorer function and $Sample_{K}(.)$ is the sampler function that produces $K$ outputs. 

The scorer ($G(.)$) comprises of multiple individual scoring modules that decide various aspects of the generated text. In our setting, the scorer intends to measure: (a) to what extent the generated text is semantically related to the input, (b) how fluent the generated text is, and, (c) how formal the text is, in terms of readability grade. These aspects are measured as follows:

\subsubsection{Semantic Relatedness:}
Semantic relatedness between the source ($\bm{X}$) and generated/sampled text ($\bm{Y}$) (denoted as $r_{s}$) indicate how well the meaning is preserved during transformation. 
$$r_{s}(\bm{X},\bm{Y}) = docsim(\bm{X},\bm{Y})$$ where $docsim(.)$ can be any document similarity measure. We use a simplistic embedding based similarity measure where a cosine similarity between averaged embedding vectors of $\bm{X}$ and $\bm{Y}$ is considered as similarity\footnote{We use Spacy's (spacy.io) document similarity function. Similarity between texts are computed after removing stopwords.}. 

\subsubsection{Fluency:}
Fluency (denoted as $r_{f}$) or how structurally well constructed is the target text ($\bm{Y}$), is measured by using a language model. 
$$r_{f}(\bm{Y}) = p(\bm{y_1},\bm{y_2},\cdots, \bm{y_{M}})$$ which can be broken down into a product of conditional probability terms. For estimating the conditional probabilities, N-gram \cite{brown1992class} or neural language models \cite{bengio2003neural} can be used. For our experiments, we trained a 4-gram back-off model with KenLM \cite{heafield2011kenlm} and Europarl \textit{monolingual} English corpus \cite{koehn2005europarl}, which contains mixed domain texts.

\subsubsection{Readability Grade:}
Readability grade score (denoted as $r_{d}$) indicates the grade level that must be acquired to understand a given text. Typically a higher grade is indicative of higher linguistic complexity. Out of several metrics for readability grade scoring available, we chose the \textit{Flesch-Kincaid} readability index \cite{kincaid1975derivation}, a simple yet popular readability metric. The metric relies on number of words in a sentence, and average number of syllables per word; it thus assesses the \textit{lexical} complexity of the text. 

The cumulative scoring function $G(.)$ is a linear combination of the individual scores\footnote{scores normalized between [0,1].} discussed above. 
\begin{equation}
G(\bm{X},\bm{Y}) = \beta_s r_s(\bm{X},\bm{Y}) +\beta_f r_f (\bm{Y}) + \beta_d r_d (\bm{Y})
\label{eq:cumreward}
\end{equation}
where weights $\beta_s$, $\beta_f$, $\beta_d$ are empirically decided. 
\subsection{Sampling Strategy:}
At each time step $i$ during generation, instead of choosing the most likely word from $\bm{y_{i}}$, one of its synonyms is randomly picked. The synonym of the word is decided by looking up a synonym list\footnote{We extract synonym lexicon using English WordNet.}, and picking one synonym randomly with a uniform probability of $1/total~number~of~synonyms$.  A sentence is formed by concatenating the words thus chosen. Since large number of combinations of synonyms are possible, one iteration of sampling is stopped when $K$ sample sentences are extracted.

Note that our sampler is highly lexicalized, \textit{i.e.,} it generates only lexical variations. The reason behind such as choice is two-fold: (a) majority of the  existing scorers for readability and document similarity assessment emphasize on the lexical aspects, and (b) generic/data-independent paraphrasers that transform text at syntax, semantic levels are elusive. However, the sampler plays an external role and  our system can be bolstered with better samplers as the state-of-the-art progresses. For instance, for obtaining more diverse samples, conditional VAEs \cite{sohn2015learning}, trained on unlabeled texts can be considered.
\subsection{Control Determination}
\label{subsubsec:ctrldetermine}
The sampler and scorer result in generation of new example output for ($\bm{X},\bm{Y_s}$). However, for training the encoder-decoder (see \textit{Exploitation},  in  Section \ref{subsec:exploit}), the system needs data in the form of ($\bm{X},\bm{Y_s},c$). The control value ($\bm{c}$) for the newly generated example is determined as follows. 
\begin{equation}
c = 
\begin{cases}
1,~\text{if~}  c_{r} < \zeta_1 \\
2,~\text{if~} \zeta_1 < c_{r} < \zeta_2 \\
3,~\text{if~} c_{r} > \zeta_2,
\end{cases}
\label{eq:ctrlcriteria}
\end{equation}
where $c_{r} = r_d(Y_{s}) / r_d(X)$. 

In our setup we allow only one control that is based on readability. For other tasks and multiple controls, appropriate criteria can be defined within the same framework.

\subsection{Control Predictor}
\label{subsec:ctrlpred}
The encoder-decoder during training is coupled with a classification module or  \textit{Control Predictor}. The role of the control predictor is to predict the expected control value of any input output pair ($\bm{X},\bm{Y}$). The control predictor  constitutes a pair of encoders on top of embedding layers that get hidden representations $\bm{h_{x}}$, $\bm{h_{y}}$ from both $\bm{X}$ and $\bm{Y}$ respectively. The hidden representations are then concatenated $\bm{h} = [\bm{h_{x}}; \bm{h_{y}}]$ and passed to a fully connected neural network with \texttt{ReLU} activation to get an intermediate hidden representation $\bm{h_{f}}$. To obtain a probability vector we perform a linear transform over $\bm{h_{f}}$ and normalize it using softmax, which is given as:
\begin{gather}
\bm{h_{f}} = \texttt{relu}~(\bm{W_{h}} \bm{h}) \\
\bm{h_c} = \texttt{softmax}~(\bm{W_{c}} \bm{h_{f}})
\end{gather}



During training, the control predictor takes $<\bm{X},\bm{Y_s},\bm{c}>$ instances generated by the sampler and scorer module. It predicts control $\bm{\hat{c}}$, suffers a \textit{cross-entropy} based loss $\mathcal{L}_{2} (\bm{c}, \bm{\hat{c}})$. The losses are back-propagated to the embedding layers. Section \ref{subsec:rationale} motivates the need of this module.
\section{Training Objective and Process}
Training is carried out in three phases: (1) \textit{pre-training} (2) \textit{exploration} and (3) \textit{exploitation}. While phase (1) is carried out once, the system undergoes phases (2) and (3) for several iterations. Figure \ref{fig:architecture} gives an overview of the training process.
\subsection{Pre-training}
The encoder-decoder is first pretrained as an autoencoder  with unlabeled data, it learns to predict $X'$ from $X$ such that $X'\sim X$. Since, no control input is available during this phase, the control input is neglected. Pretraining ensures better initialization of the encoder and decoder parameters.
\subsection{Exploration}
\label{subsec:explore}
Since labeled data is not directly available for training, the exploration phase helps synthesize instances of input text, output text and appropriate control ($<\bm{X},\bm{Y},\bm{c}>$). The system is not trained during the process and is only allowed to predict output $\bm{Y_g}$ and ``explore'' for other samples that are variations of $\bm{Y_g}$. With the help of the sampler, the system generates $K$ sample outputs and the scorer selects the sample $\bm{Y_s}$ that maximizes the score. If none of the samples obtain a better score than the default output $\bm{Y_g}$, no new instance is augmented for that particular $X$ in that particular iteration of exploration. If the generated output is same as the input, that instance is discarded. The possible control value for the generated sample $\bm{Y_s}$ is computed following Equation \ref{eq:ctrlcriteria}.

As the encoder-decoder is pretrained, in the first iteration of exploration, the model predicts $\bm{Y_g}$ same as input $\bm{X}$. Since the sampled sentences selected by the scorer will always have better score than $\bm{Y_g}$ (else it will not be selected), the first iteration of exploration ensures that the output side of the synthesized data is different and with a better cumulative score than the input.
\subsection{Exploitation}
\label{subsec:exploit}
In this phase, the encoder-decoder are trained using the data generated during the exploration phase. This process is carried out in two stages as mentioned below:
\subsubsection{Training of Control Predictor:}
The Control Predictor Module is trained with $\bm{X},\bm{Y}$ taken as input and $\bm{c}$ as output. It undergoes training in a standard classification setup where batches of labeled data are fed in multiple iterations and the loss is minimized. Once training gets over, the predictor is plugged into the encoder-decoder network (where it becomes non-trainable) to predict the control for the generated sentences. Both $\bm{X},\bm{Y}$ are provided as input since control values signify the \textbf{relative} variation of $\bm{Y}$ w.r.t. $\bm{X}$. 

\subsubsection{Training of Encoder-Decoder:}
The encoder-decoder framework trains with source $\bm{X}$, control $\bm{c}$ as input and target output $\bm{Y_{s}}$. For each instance, the model predicts output $\bm{Y_{g}}$. The control predictor then has to predict the control category of the output ($\bm{Y_{g}}$). However, obtaining $\bm{Y_{g}}$ involves finding out the most likely words in the conditional probability distribution (Equation \ref{eq:cond}) using \texttt{argmax} operation. This makes the overall loss (Equation \ref{eq:composite}) non-differentiable.  

To avoid this, we approximate $Y_g$ as follows:
\begin{equation*}
Y_g \approx \Tilde{Y}_{g} = \{\Tilde{y}_1,\Tilde{y}_2,...,\Tilde{y}_{M}\}
\end{equation*}
where, 
\begin{equation*}
	\Tilde{y}_{i} \sim \texttt{softmax}(s_{i} / \tau)
\end{equation*}
where, $s_{i}$ is the unnormalized decoder hidden representation (logits) and $\tau > 0$ is the \textit{temperature}. Setting the temperature to close to zero approximates the output to one-hot representation, typically obtained through \texttt{argmax}. This, however, ensures differentiability unlike \texttt{argmax}. 

The modified output $\bm{\Tilde{Y}_{g}}$ is input to the Control Predictor which predicts the appropriate control $\bm{\hat{c}}$ pertaining to $\bm{X}$ and $\bm{Y_{g}}$. Based on the predicted output and control, a composite loss is calculated, as follows. 
\begin{equation}
\begin{split}
\mathcal{L}(\bm{Y_{s}},\bm{Y_{g}},\bm{c},\bm{\hat{c}}) = \lambda \times \mathcal{L}_{2}(\bm{c},\bm{\hat{c}}) \\
+ (1-\lambda)\times \mathcal{L}_{1}(\bm{Y_s},\bm{\Tilde{Y}_g})
\end{split}
\label{eq:composite}
\end{equation}
where $\lambda \in [0,1]$ is a hyper parameter and the functions $\mathcal{L}_{1} (.)$ and $\mathcal{L}_{2} (.)$ are \textit{cross-entropy} based loss functions, popularly used in sequence-to-sequence  and classification tasks.

\subsection{Why Control Predictor?}
\label{subsec:rationale}
In our setting for transformation, token level entropy faces the following issues: (i) Token-level loss averaged over the generated words equally penalizes all words; but in controllable transformation some words (for instance  \textit{content} words) play a more important role than others (\textit{e.g.} \textit{stop-words}), (ii) Cross entropy alone does not adequately capture how the generated sentences are related to the controls passed. To alleviate these, we employ a predictor that computes an additional loss between predicted controls of the generated sentence and the reference controls.
\section{Experiment Setup}
\label{sec:setup}

We now describe the dataset, architectural choices, and hyperparameters selected. Our dataset contains \textbf{unlabeled} text which are simple and informal in nature. It comprises of $14432$ informally written sentences  collected from \textit{Enron Email Corpus}\footnote{\url{http://bailando.sims.berkeley.edu/enron_email.html}}, \textit{Corpus of Late Modern English Prose}\footnote{\url{http://ota.ox.ac.uk/desc/2077}},\textit{ non-spam emails from Spam Dataset}\footnote{\url{http://csmining.org/index.php/spam-assassin-datasets.html}}, and \textit{essays for kids}\footnote{\url{http://www.mykidsway.com/essays/}}. The data is split into a train:valid:test split of $80\%:12\%:8\%$. The vocabulary size of the dataset is $95775$ words. The source code and dataset are available at \url{https://github.com/parajain/uctf}.



We used a bidirectional GRU based encoder with 2 layers; the forward and backward hidden representations from the encoder are concatenated. The embedding matrix with a word embedding size of 300 was shared between the encoder and decoder. For encoder, 2 layers of GRUs were used and the hidden dimension size was set to 250. The decoder had 2 layers with hidden dimension size set to 500 . The weights related to language model, sentence similarity and readability scorers, given in Equation \ref{eq:cumreward} are empirically set to 0.6, 0.2 and 0.2 respectively. Note that, Equation \ref{eq:cumreward}, by design, allows choosing different combinations of transformation parameters. These parameters may be tuned differently for different practical settings. For example, a low weight on semantic relatedness will be less restrictive on ``relatedness'' and can generate text with higher readability grade which is suitable for a domain such as creative writing. On the other hand for a domain such as legal where relatedness is more important, the corresponding weight can be set to a higher value.

During loss computation (Equation \ref{eq:composite}) we tried different values of parameter $\lambda \in [0.1, 0.3, 0.5]$ and settled with $0.1$. The $\zeta_1$ and $\zeta_2$ in equation \ref{eq:ctrlcriteria} were set to $1.05$ and $1.1$ respectively; these threshold values were incremented by $5\%$ for each category to suit practical purposes. The temperature parameter $\tau$ was set to $0.001$; we did not use annealing. For both encoder-decoder and control predictor, we employed the Adam optimizer with the learning rate set to $0.001$. The system underwent pretraining once, followed by a \textit{maximum} of $20$ cycles of exploration-exploitation. For each input, we chose max $K=100$ samples. Exploitation occurs for $20$ epochs with early stopping enabled. After each cycle, the model was saved and we chose the model which provided the best average score over a held-out set prepared only for model selection. During testing, for each input sentence in the test-data, control level values of $2$ and $3$ were chosen.

\noindent \textbf{Test scenario:} The present system is configured to take an input sentence and \textbf{three} possible control levels pertaining to the degree of formalization desired by the user. The control levels can be 1, 2 and 3. As depicted in Figure \ref{fig:example} in the introductory section, an input control of 1 would retain the input sentence as is, control 2 would transform it to a \textit{relatively} medium level formal text (termed as \textit{Formalness-Mid}) and control of 3 will transform the input to a text that is even more formal that $2$, (termed as \textit{Formalness-High}). 
\begin{table*}[t]
\begin{center}
\begin{tabular}{@{}lcccccccc@{}}
\toprule
Mode &  \multicolumn{2}{c}{\textsc{Ctrl}} & \multicolumn{2}{c}{\textsc{Ctrl}}&\multicolumn{2}{c}{\textsc{Ctrl}}&\multicolumn{1}{c}{Mueller} \\
 &  \multicolumn{2}{c}{\textsc{WithPredictor}} & \multicolumn{2}{c}{\textsc{NoPredictor}}&\multicolumn{2}{c}{\textsc{OneShot}} & \multicolumn{1}{c}{et al., 2017}\\
 \toprule
{Formalness} &&&&&&&\\
{Control}&{Mid} &{High}&{Mid}&{High}&{Mid} &{High} & NONE \\
\midrule
Readability &   0.568  &  0.583   & 0.538  &  0.538  & 0.554  & 0.554 & 0.33  \\
Relatedness & 0.72  & 0.74   & 0.77  & 0.77   & 0.78  & 0.78 & 0.05  \\
LM Score & 0.34  & 0.34  & 0.32  & 0.32  & 0.30 & 0.30 & 0.16 \\
\bottomrule
\end{tabular}
\end{center}
\caption{Average test-set scores (normalized between $[0-1]$)} 
\label{tab:allscores}
\end{table*}
\begin{table}[t]
\begin{center}
\footnotesize
\setlength\tabcolsep{0.1cm}
\begin{tabular}{c c c | c}
\toprule
System &  BLEU  & Relatedness & Readability\\
& & (with input) & \\
\midrule
Mueller et al. & 5.09  & 0.41 & 0.38\\
Seq2Seq (skyline) & 38.37 & 0.17 & 0.71 \\
Formalness-Mid (Ours) & 21.81  & 0.58 &  0.52\\
Formalness-High (Ours) & 21.14  & 0.57 & 0.74 \\
\bottomrule
\end{tabular}
\end{center}
\caption{Comparison of BLEU (\%), avg. Semantic Relatedness with input, and  avg. normalized readability scores. \textbf{The avg. readability for input and reference sentences in test data are (0.41) and (0.50) respectively}}
\label{tab:bleu}
\end{table}
\subsection{Research Questions Addressed}
Our experiments are designed to address the following research questions
\begin{itemize}
\item \textbf{RQ1:} Is our method able to produce output that is formal, fluent and adequate? Is the transformation controllable? How well does the output's formalness conform to the input control provided?
\item \textbf{RQ2:} Does the control predictor have a positive impact on the overall performance of our model? 
\item \textbf{RQ3:} Is iterative exploration-exploitation helpful? Why wouldn't a much simpler strategy of just one round of sampling and training suffice?
\item \textbf{RQ4:} Does our approach outperform the existing supervised and unsupervised methods for such tasks? 
\end{itemize}

\textbf{RQ1} is answered by measuring fluency, semantic relatedness and readability of the output text following Section \ref{subsec:sampler}. To answer \textbf{RQ2} and \textbf{RQ3}, we prepare two trivial variants of system: (a) \textsc{CtrlNoPredictor}, in which the control predictor module is removed, and (b) \textsc{CtrlOneShot} where exploration and exploitation cycles are not iteratively carried out. Rather, the system undergoes one round of exploration cycles followed by one exploitation cycle. 

For \textbf{RQ4}, since the existing approaches/systems have no provisions to take control parameters like ours, comparison is only done based on their transformation capability.
\begin{itemize}
\item \textbf{Comparison with Existing Unsupervised Method:} The most relevant unsupervised transformation system is by \newcite{mueller2017sequence}. This system takes a sequence as input, auto-encodes it and applies minor variation based on expected outcome that is decided by a scorer. We fix readability metric as the scoring function and treat the system as a baseline.
\item \textbf{Indirect Comparison with Supervised Systems}
A similar task to ours is the reverse of text  simplification, for which labeled datasets exist. We chose the dataset configuration given in the neural-text simplification work by \newcite{nisioi2017exploring}, originally derived from \newcite{hwang2015aligning}. We flip the input-output side of the parallel corpus, thereby reversing the training objective. Our system uses only the simplified side of the data for training, with the previously discussed configuration. Additionally, the complete \emph{simple-to-complex} parallel corpus was used to train a neural Seq2Seq generation system \cite{bahdanau2014neural} using the Marian toolkit \cite{junczys2018marian} with default configurations\footnote{convergence attained after $70000$ iterations.}.
\end{itemize}
\begin{figure}[t]
\centering
\includegraphics[width=.75\columnwidth, keepaspectratio]{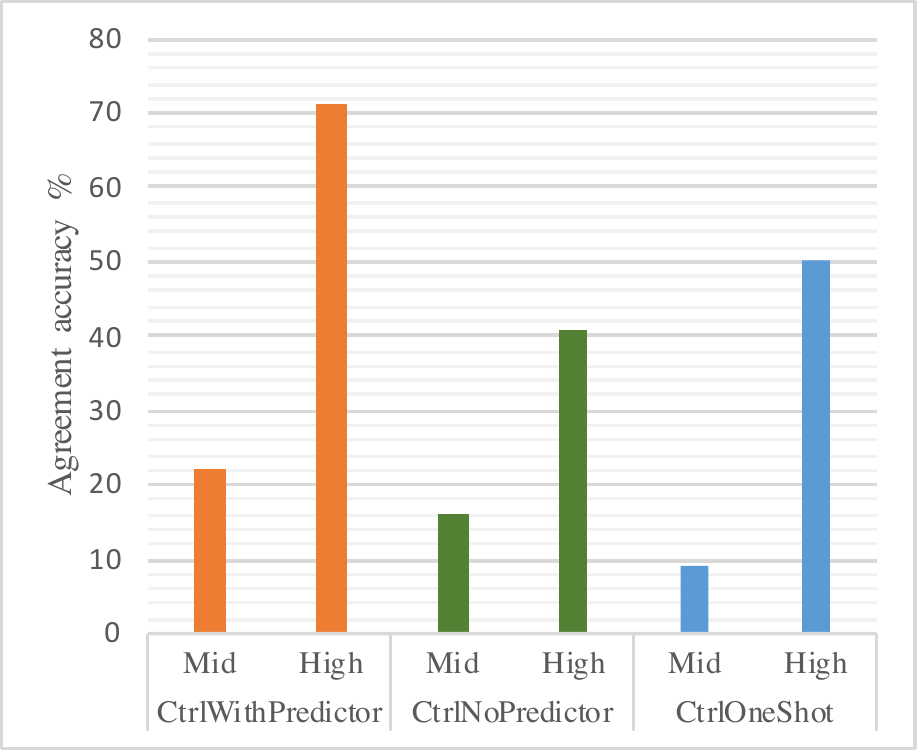}
\caption{Accuracy (in \%) showing the agreement between desired input control and control measured on the output text using Equation \ref{eq:ctrlcriteria}}. 
\label{fig:agreement}
\end{figure}

\begin{table*}
\centering
\small
\setlength\tabcolsep{0.1cm}
\begin{tabular}{>{\arraybackslash}m{0.8cm}>{\arraybackslash}m{4.8cm}>{\arraybackslash}m{4.8cm}>{\arraybackslash}m{4.8cm}}
\midrule
Mode &  {Input Sentence } &{\textit{Formalness Control-Mid}}&{\textit{Formalness Control-High}}\\
\midrule
 \multirow{4}{*}{\STAB{\rotatebox[origin=l]{90}{WithPred}}}
&(1) 18 year old who abandoned her child in a hospital later got custody &  (1) 18 year old who unpopulated her kid in a infirmary resultant got custody
& (1) 18 year old who deserted her tyke in a infirmary resultant got detention \\
   &(2) the first sync after upgrading will be slow & (2) the first synchronise afterward upgrading will be idle &  (2) the first synchronise afterward upgrading will be laggard \\  
&&&\\
\hline 
&&&\\
 \multirow{4}{*}{\STAB{\rotatebox[origin=l]{90}{NoPred}}}
   &(1) 18 year old who abandoned her child in a hospital later got custody
&(1) 18 class old who deserted her child in a infirmary accompanying got detention
&(1) 18 class old who deserted her child in a infirmary accompanying got detention
 \\ 
 
 &(2) the first sync after upgrading will be slow
&(2) the introductory synchronise afterward upgrading will be goosy 
&(2) the introductory synchronise afterward upgrading will be goosy 
\\
&&&\\
\hline
&&&\\
 \multirow{4}{*}{\STAB{\rotatebox[origin=l]{90}{OneShot}}}
   &(1) 18 year old who abandoned her child in a hospital later got custody
&(1) 18 yr old who untenanted her tyke in a hospital subsequently got detention
&(1) 18 class old who deserted her tyke in a hospital subsequently got detention \\ 
&(2) the first sync after upgrading will be slow
&(2) the eightieth sync later upgrading bequeath be tedious
&(2) the eightieth sync later upgrading bequeath be tedious
\\
&&&\\
\hline
\end{tabular}
\caption{Example input and transformed sentences with varying control values and system variants}
\label{tab:example}
\end{table*}
\section{Evaluation Results}
\label{sec:results}
For our dataset, we first evaluate the output text based on: (a) basic parameters such as fluency, adequacy (semantic relatedness with input) and readability grade, and (b) whether the quality of output comply with the input control parameters or not. A direct comparison between different variants of our own system could be done using these measures. 

Observe in Table \ref{tab:allscores}, the readability scores for \textsc{CtrlWithPredictor} are always better than the average readability grade of the input (\bm{$0.54$}). This indicates that the transformation obtained by the system has an improved readability grade. The improvement is even more for \textsc{CtrlWithPredictor} than the other two, which demnostrates the importance of the Control Predictor and the iterative training scheme. The Language Model scores \textsc{CtrlWithPredictor} is the highest and differs significantly from \textsc{CtrlNoPredictor}. This shows that adding the auxiliary loss from Control Predictor output results in better fluency. Regarding semantic relatedness, we observe that most sentences generated by \textsc{CtrlOneShot} undergo little or no modification \textit{vis-\'a-vis} the input. This results in a slightly higher semantic relatedness score. Moreover, in most cases the output obtained from these two systems are indifferent to control values of $Mid$ and $High$, hence the average scores are same for both control levels. For \textsc{CtrlOneShot}, for the same input text, multiple variants for the same input are generated in one-shot. In the hindsight, using such data may have confused the system during training and it may have learned to ignore the control. Lastly, \textsc{CtrlNoPredictor} hardly incurs any direct loss pertaining to the control values, and thus, becomes somewhat agnostic of the control levels.

Figure \ref{fig:agreement} indicates how robustly the systems respond to the input control. The agreement accuracy in the figure refers to the percent of test-instances for which the output control (measured using Equation \ref{eq:ctrlcriteria}) match the intended input control. For all the three systems, for control value $Mid$, most of the generated text actually are either confused with control value $Default$ (\textit{i.e.},very similar to input) or control value $High$. This suggests the need for fine-tuning the user-defined parameters in Equation \ref{eq:ctrlcriteria}. However, note that the highest percentage values are obtained for \textsc{CtrlWithPredictor}. 
\subsection{Human Evaluation}
We performed an additional experiment where three language experts unaware of the task were given 30 random instances from our test dataset. Each instance consists of the input sentence, the predicted outputs with controls $Mid$ and $High$ (shuffled for ruling out bias). For each instance the experts were asked to rank the output sentences based on their perceived readability. Ideally, the output for $Mid$ should rank lower than the one for $High$. The average agreement between the human-rated rank labels and the default ranking based on input control categories turned out to be \textbf{80.2\%}, This indicates that the readability of output $High$ is indeed higher than that of $Mid$.

Table \ref{tab:example} presents a few randomly selected examples for different input sentences and varied control values for all 3 system variants. We can clearly observe that the input sentences are modified according to control values. \textsc{CtrlWithPredictor} shows increasing readability grade going from $Mid$ to $High$.  This is also corroborated by readability scores in Table \ref{tab:allscores}. Further, as expected, the \textsc{CtrlNoPredictor} system is unable to capture the change in control levels due to lack of explicit feedback. The \textsc{CtrlOneShot} system also does not show much improvement with controls due to lack of transformation knowledge it would have gained if the iterative exploration/exploitation based training had been applied. We believe, our observations above provide positive answers to research questions \textbf{RQ1}, \textbf{RQ2}, and \textbf{RQ3}. 

\subsection{Comparison with Supervised Systems for Reversed-simplification Task}
Table \ref{tab:bleu} shows how comparable are the generated complex sentences with the reference complex texts for reverse simplification task. For this, BLEU \cite{papineni2002bleu} is considered as the metric. Supervised Seq2Seq, as expected serves as a \textit{skyline} and achieves a much better BLEU than ours. Yet the BLEU score of our system is significant, considering that our system is unsupervised and not designed to produce outputs that overlap more with the reference text. It is important to note that the average readability scores for our system variants are better than that of the reference corpora (avg. readability 0.50) and Seq2Seq, which shows the capability of our system in transforming the input to a more formal version. Moreover, the higher semantic relatedness scores indicate that our models are indeed capable of preserving the semantics better.

Regarding comparison with Mueller et al., the crux of their system is a variational autoencoder and an outcome prediction module which revises the input such that the output has a higher expected outcome. The problem with such an approach is that it is restricted to generating from a known distribution of sentences provided in the form of training data. The reduced performance by the system as seen in Tables \ref{tab:allscores} and \ref{tab:bleu} is perhaps due to the facts that the training data size is less and the system does not have schemes for iterative training like ours. The above comparison studies provide insights for \textbf{RQ4}.
\section{Conclusion and Future Work}
\label{sec:concl}
We proposed a novel NLG framework for unsupervised controllable text transformation that relies on \textit{off-the-shelf} language processing systems for acquiring transformation knowledge. Our system is tested for the task of formalizing the input text by improving its readability grade, where the degree of readability grade is controllable. Experiments on general domain datasets demonstrate the goodness of the output transformed texts both quantitatively and qualitatively. The system also learns to give importance to the user-defined control and responds accordingly. A shortcoming of the current approach is that it only performs lexical formalization because of the sampling strategy and choice of Flesch Kincaid readability metric that is highly lexical. Our future agenda involves exploring better sampling techniques for generating complex structural and semantic variants, and also testing the framework for tasks like text simplification and style-transformation.
\bibliographystyle{aaai}
\bibliography{tunable}  

\begin{thebibliography}{}

\bibitem[\protect\citeauthoryear{Artetxe \bgroup et al\mbox.\egroup
  }{2017}]{artetxe2017unsupervised}
Artetxe, M.; Labaka, G.; Agirre, E.; and Cho, K.
\newblock 2017.
\newblock Unsupervised neural machine translation.
\newblock {\em arXiv:1710.11041}.

\bibitem[\protect\citeauthoryear{Bahdanau, Cho, and
  Bengio}{2014}]{bahdanau2014neural}
Bahdanau, D.; Cho, K.; and Bengio, Y.
\newblock 2014.
\newblock Neural machine translation by jointly learning to align and
  translate.
\newblock {\em arXiv:1409.0473}.

\bibitem[\protect\citeauthoryear{Bengio \bgroup et al\mbox.\egroup
  }{2003}]{bengio2003neural}
Bengio, Y.; Ducharme, R.; Vincent, P.; and Jauvin, C.
\newblock 2003.
\newblock A neural probabilistic language model.
\newblock {\em JMLR, 2003} 3(Feb):1137--1155.

\bibitem[\protect\citeauthoryear{Brown \bgroup et al\mbox.\egroup
  }{1992}]{brown1992class}
Brown, P.~F.; Desouza, P.~V.; Mercer, R.~L.; Pietra, V. J.~D.; and Lai, J.~C.
\newblock 1992.
\newblock Class-based n-gram models of natural language.
\newblock {\em Computational linguistics}.

\bibitem[\protect\citeauthoryear{Cho \bgroup et al\mbox.\egroup
  }{2014}]{cho2014properties}
Cho, K.; Van~Merri{\"e}nboer, B.; Bahdanau, D.; and Bengio, Y.
\newblock 2014.
\newblock On the properties of neural machine translation: Encoder-decoder
  approaches.
\newblock {\em arXiv:1409.1259}.

\bibitem[\protect\citeauthoryear{Danescu-Niculescu-Mizil \bgroup et
  al\mbox.\egroup }{2013}]{danescuniculescumizil-EtAl:2013:ACL2013}
Danescu-Niculescu-Mizil, C.; Sudhof, M.; Jurafsky, D.; Leskovec, J.; and Potts,
  C.
\newblock 2013.
\newblock A computational approach to politeness with application to social
  factors.
\newblock In {\em ACL, 2013}.

\bibitem[\protect\citeauthoryear{Ficler and
  Goldberg}{2017}]{ficler2017controlling}
Ficler, J., and Goldberg, Y.
\newblock 2017.
\newblock Controlling linguistic style aspects in neural language generation.
\newblock {\em arXiv:1707.02633}.

\bibitem[\protect\citeauthoryear{Gatt and Reiter}{2009}]{gatt2009simplenlg}
Gatt, A., and Reiter, E.
\newblock 2009.
\newblock Simplenlg: A realisation engine for practical applications.
\newblock In {\em 12th European Workshop on NLG}.

\bibitem[\protect\citeauthoryear{Heafield}{2011}]{heafield2011kenlm}
Heafield, K.
\newblock 2011.
\newblock Kenlm: Faster and smaller language model queries.
\newblock In {\em Sixth Workshop on SMT}.

\bibitem[\protect\citeauthoryear{Hu \bgroup et al\mbox.\egroup
  }{2017}]{hu2017toward}
Hu, Z.; Yang, Z.; Liang, X.; Salakhutdinov, R.; and Xing, E.~P.
\newblock 2017.
\newblock Toward controlled generation of text.
\newblock In {\em International Conference on Machine Learning}.

\bibitem[\protect\citeauthoryear{Hwang \bgroup et al\mbox.\egroup
  }{2015}]{hwang2015aligning}
Hwang, W.; Hajishirzi, H.; Ostendorf, M.; and Wu, W.
\newblock 2015.
\newblock Aligning sentences from standard wikipedia to simple wikipedia.
\newblock In {\em NAACL-HLT}.

\bibitem[\protect\citeauthoryear{Jain \bgroup et al\mbox.\egroup
  }{2018}]{N18-2098}
Jain, P.; Laha, A.; Sankaranarayanan, K.; Nema, P.; Khapra, M.~M.; and Shetty,
  S.
\newblock 2018.
\newblock A mixed hierarchical attention based encoder-decoder approach for
  standard table summarization.
\newblock In {\em NAACL-HLT}.

\bibitem[\protect\citeauthoryear{Junczys-Dowmunt \bgroup et al\mbox.\egroup
  }{2018}]{junczys2018marian}
Junczys-Dowmunt, M.; Grundkiewicz, R.; Dwojak, T.; Hoang, H.; Heafield, K.;
  Neckermann, T.; Seide, F.; Germann, U.; Aji, A.~F.; Bogoychev, N.; Martins,
  A. F.~T.; and Birch, A.
\newblock 2018.
\newblock Marian: Fast neural machine translation in c++.
\newblock {\em arXiv:1804.00344}.

\bibitem[\protect\citeauthoryear{Kincaid \bgroup et al\mbox.\egroup
  }{1975}]{kincaid1975derivation}
Kincaid, J.~P.; Fishburne~Jr, R.~P.; Rogers, R.~L.; and Chissom, B.~S.
\newblock 1975.
\newblock Derivation of new readability formulas (automated readability index,
  fog count and flesch reading ease formula) for navy enlisted personnel.
\newblock Technical report, DTIC Document.

\bibitem[\protect\citeauthoryear{Koehn}{2005}]{koehn2005europarl}
Koehn, P.
\newblock 2005.
\newblock Europarl: A parallel corpus for statistical machine translation.
\newblock In {\em MT summit}.

\bibitem[\protect\citeauthoryear{Koehn}{2009}]{koehn2009statistical}
Koehn, P.
\newblock 2009.
\newblock {\em Statistical machine translation}.
\newblock Cambridge University Press.

\bibitem[\protect\citeauthoryear{Lample, Denoyer, and
  Ranzato}{2017}]{lample2017unsupervised}
Lample, G.; Denoyer, L.; and Ranzato, M.
\newblock 2017.
\newblock Unsupervised machine translation using monolingual corpora only.
\newblock {\em arXiv:1711.00043}.

\bibitem[\protect\citeauthoryear{Li \bgroup et al\mbox.\egroup
  }{2017}]{li2017paraphrase}
Li, Z.; Jiang, X.; Shang, L.; and Li, H.
\newblock 2017.
\newblock Paraphrase generation with deep reinforcement learning.
\newblock {\em arXiv:1711.00279}.

\bibitem[\protect\citeauthoryear{Mueller, Gifford, and
  Jaakkola}{2017}]{mueller2017sequence}
Mueller, J.; Gifford, D.; and Jaakkola, T.
\newblock 2017.
\newblock Sequence to better sequence: continuous revision of combinatorial
  structures.
\newblock In {\em ICML, 2017},  2536--2544.

\bibitem[\protect\citeauthoryear{Nisioi \bgroup et al\mbox.\egroup
  }{2017}]{nisioi2017exploring}
Nisioi, S.; {\v{S}}tajner, S.; Ponzetto, S.~P.; and Dinu, L.~P.
\newblock 2017.
\newblock Exploring neural text simplification models.
\newblock In {\em ACL, 2017}.

\bibitem[\protect\citeauthoryear{Niu and Bansal}{2018}]{niu2018polite}
Niu, T., and Bansal, M.
\newblock 2018.
\newblock Polite dialogue generation without parallel data.
\newblock {\em arXiv:1805.03162}.

\bibitem[\protect\citeauthoryear{Papineni \bgroup et al\mbox.\egroup
  }{2002}]{papineni2002bleu}
Papineni, K.; Roukos, S.; Ward, T.; and Zhu, W.-J.
\newblock 2002.
\newblock Bleu: a method for automatic evaluation of machine translation.
\newblock In {\em ACL, 2002}.

\bibitem[\protect\citeauthoryear{Prabhumoye \bgroup et al\mbox.\egroup
  }{2018}]{prabhumoye2018style}
Prabhumoye, S.; Tsvetkov, Y.; Salakhutdinov, R.; and Black, A.~W.
\newblock 2018.
\newblock Style transfer through back-translation.
\newblock {\em arXiv:1804.09000}.

\bibitem[\protect\citeauthoryear{Prakash \bgroup et al\mbox.\egroup
  }{2016}]{prakash2016neural}
Prakash, A.; Hasan, S.~A.; Lee, K.; Datla, V.; Qadir, A.; Liu, J.; and Farri,
  O.
\newblock 2016.
\newblock Neural paraphrase generation with stacked residual lstm networks.
\newblock {\em arXiv:1610.03098}.

\bibitem[\protect\citeauthoryear{Sheika and Inkpen}{2012}]{seika2012classify}
Sheika, F.~A., and Inkpen, D.
\newblock 2012.
\newblock Learning to classify documents according to formal and informal
  style.
\newblock {\em Linguistic Issues in Language Technology}.

\bibitem[\protect\citeauthoryear{Sheikha and
  Inkpen}{2011}]{sheikha2011generation}
Sheikha, F.~A., and Inkpen, D.
\newblock 2011.
\newblock Generation of formal and informal sentences.
\newblock In {\em 13th European Workshop on NLG}.

\bibitem[\protect\citeauthoryear{Shen \bgroup et al\mbox.\egroup
  }{2017}]{shen2017style}
Shen, T.; Lei, T.; Barzilay, R.; and Jaakkola, T.
\newblock 2017.
\newblock Style transfer from non-parallel text by cross-alignment.
\newblock In {\em NIPS, 2017},  6830--6841.

\bibitem[\protect\citeauthoryear{Sohn, Lee, and Yan}{2015}]{sohn2015learning}
Sohn, K.; Lee, H.; and Yan, X.
\newblock 2015.
\newblock Learning structured output representation using deep conditional
  generative models.
\newblock In {\em NIPS}.

\bibitem[\protect\citeauthoryear{Sutskever, Vinyals, and
  Le}{2014}]{sutskever2014sequence}
Sutskever, I.; Vinyals, O.; and Le, Q.~V.
\newblock 2014.
\newblock Sequence to sequence learning with neural networks.
\newblock In {\em NIPS}.

\bibitem[\protect\citeauthoryear{Wubben, Van Den~Bosch, and
  Krahmer}{2010}]{wubben2010paraphrase}
Wubben, S.; Van Den~Bosch, A.; and Krahmer, E.
\newblock 2010.
\newblock Paraphrase generation as monolingual translation: Data and
  evaluation.
\newblock In {\em INLG, 2010}.

\bibitem[\protect\citeauthoryear{Yu \bgroup et al\mbox.\egroup
  }{2017}]{Yu2017SeqGANSG}
Yu, L.; Zhang, W.; Wang, J.; and Yu, Y.
\newblock 2017.
\newblock Seqgan: Sequence generative adversarial nets with policy gradient.
\newblock In {\em AAAI, 2017}.

\end{thebibliography}
\end{document}